\documentclass{article}
\usepackage{spconf,amsmath,graphicx}
\usepackage{times}
\usepackage{epsfig}
\usepackage{graphicx}
\usepackage{amsmath}
\usepackage{amssymb}
\usepackage{booktabs}
\usepackage{comment}

\usepackage[ruled,vlined]{algorithm2e}

\usepackage{amsthm}
\usepackage{subcaption}
\usepackage{multirow}
\usepackage{makecell}

\title{TransAdapt: A Transformative Framework for Online Test Time Adaptive 
Semantic Segmentation}
\name{\small {Debasmit Das, Shubhankar Borse, Hyojin Park, Kambiz Azarian, {Hong Cai}, {Risheek Garrepalli}, {Fatih Porikli}}}
\address{Qualcomm AI Research*\thanks{*Qualcomm AI Research is an initiative of Qualcomm Technologies, Inc.}\\
\tt\small \{debadas, sborse, hyojinp, kambiza, hongcai, rgarrepa, fporikli\}@qti.qualcomm.com}

\begin{document}
\ninept
\maketitle
\begin{abstract}
Test-time adaptive (TTA) semantic segmentation adapts a source pre-trained image semantic segmentation model to unlabeled batches of target domain test images, different from real-world, where samples arrive one-by-one in an online fashion. To tackle online settings, we propose TransAdapt, a framework that uses transformer and input transformations to improve segmentation performance. Specifically, we pre-train a transformer-based module on a segmentation network that transforms unsupervised segmentation output to a more reliable supervised output, without requiring test-time online training. To also facilitate test-time adaptation, we propose an unsupervised loss based on the transformed input that enforces the model to be invariant and equivariant to photometric and geometric perturbations, respectively. Overall, our framework produces higher quality segmentation masks with up to 17.6\% and 2.8\% mIOU improvement over no-adaptation and competitive baselines, respectively. 
\end{abstract}

\section{Introduction}


Deep learning systems produce highly accurate predictions when tested on data similar to the training data. However, when there is a distribution shift between training and test data, the performance of deep learning systems can be significantly impacted. In particular, previous work in semantic segmentation, which is a key computer vision task for various applications like self-driving and AR/VR, has often seen such performance degradation caused by domain gaps. More specifically, researchers and practitioners often utilize synthetic data~\cite{RichterECCV16PlayingForData,RosCVPR16SYNTHIADataset} to train semantic segmentation models, since obtaining ground-truth annotations on real images is very costly. However, such trained models usually perform poorly on real images due to the drastic visual difference between synthetic and real data.


          



In order to reduce the gap, researchers have proposed various domain adaptation approaches that include self-training with pseudo-labels~\cite{LiCVPR19BidirectionalLearningDASemSegm,ZouECCV18UnsupervisedDASemSegmClassBalancedSelfTraining,ChenICCV17NoMoreDiscriminationCrossCityAdaptationRoadSceneSegmenters,DuICCV19SSFDANSeparatedSemanticFeatureDASemSegm,zhang2022perceptual,borse2021hs3,VuCVPR19ADVENTAdversarialEntropyMinimizationDASemSegm}, adversarial feature alignment~\cite{HoffmanX16FCNsInTheWildPixelLevelAdversarialDA,HuangECCV18DomainTransferThroughDeepActivationMatching,ZhangCVPR20TransferringRegularizingPredictionSemSegm}, input style transfer~\cite{HoffmanICML18CyCADACycleConsistentAdversarialDA,MurezCVPR18ImageToImageTranslationDA,SankaranarayananCVPR18LearningSyntheticDataDomainShiftSemSegm,WuECCV18DCANDualChannelWiseAlignmentNetworksUDA,ChangCVPR19AllAboutStructureDASemSeg,YangCVPR20PhaseConsistentEcologicalDA,ChenCVPR19CrDoCoPixelLevelDomainTransferCrossDomainConsistency}, or conditioning of segmentation outputs~\cite{TsaiCVPR18LearningToAdaptStructuredOutputSpaceSemSegm,VuCVPR19ADVENTAdversarialEntropyMinimizationDASemSegm,PanCVPR20UnsupervisedIntraDASemSegmSelfSupervision}. 
 Under the assumption that a large number of unlabeled images are available from the target/test domain, one can finetune the pretrained semantic segmentation model on both the unlabeled test-domain data (via an unsupervised loss) as well as the labeled source-domain data.  This produces a domain-invariant model, which can produce more accurate predictions on the target domain as compared to the pretrained model.

{There exists source-free domain adaptation methods~\cite{WangICLR21TENTFullyTestTimeAdaptEntropyMin,SunICML20TestTimeTrainingSelfSupervisionGeneralizDistr,SchneiderNeurIPS2020ImprovingRobustnessCommonCorruptionsCovShiftAdapt, choi2022improving, azarian2023test} that assume absence of source domain data during adaptation. However, these methods can lead to improved segmentation performance on common evaluation benchmarks due to adaptation on large target data batches, which can  overestimate segmentation performance.} Specifically, in many real-world deployments, prior access to a set of unlabeled target-domain data will not be available for performing offline model updates. On the contrary, the target-domain data samples often arrive on-the-fly. 
Most existing domain adaptation methods can not be used in this online setting, since gradient updates based on single images will be noisy and degrade the model's stability and accuracy.

\begin{figure*}[h]
\centering
\includegraphics[width=0.98\linewidth]{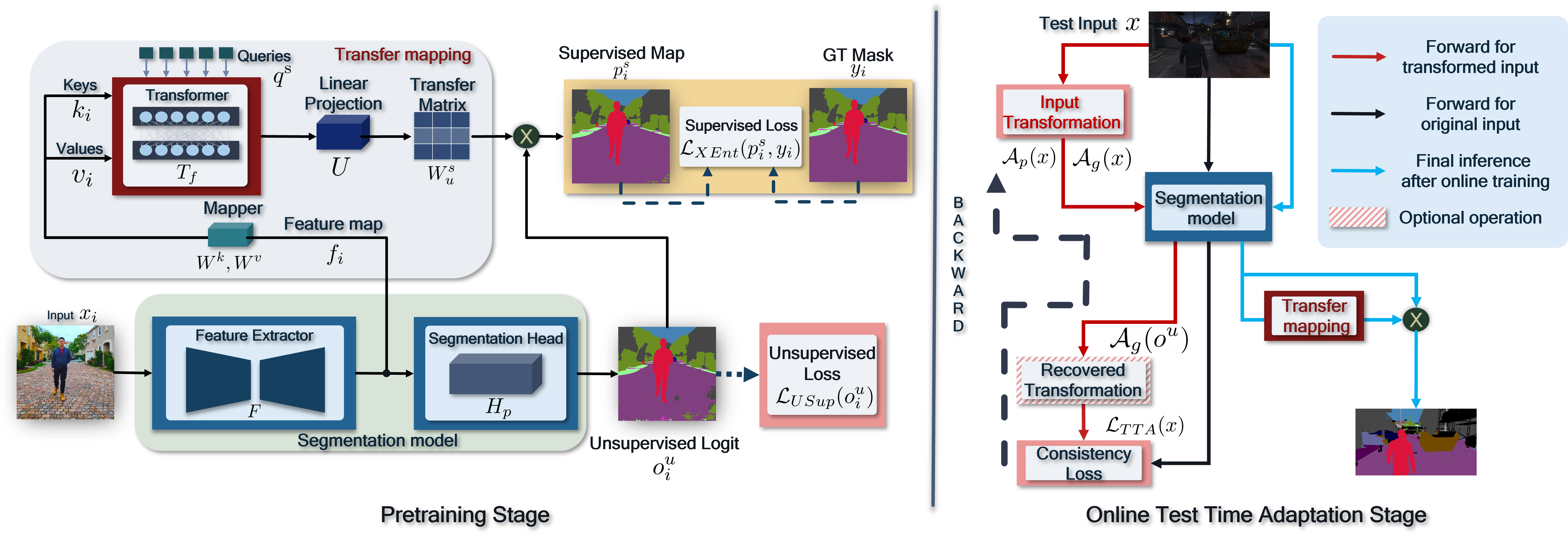}
\vspace{-5pt}
\caption{\small The feature extractor, prediction head, transformer module and other learnable parameters are pre-trained using a combination of supervised and unsupervised losses. During test time, the unsupervised loss is used to conduct adaptation. Once the adaptation is done, we use the output from the prediction head and multiply it with the transfer matrix to produce the supervised output to for inference.}
\label{fig:pretrain}
\vspace{-15pt}
\end{figure*}

In this paper, we formally introduce online test-time adaptive semantic segmentation, where we adapt a pre-trained model on online test image sequences without accessing source domain data. On this task, we construct three cross-dataset benchmarks by evaluating existing domain adaptation methods to establish several baselines. In addition to establishing baselines, we propose a novel framework, namely \emph{TransAdapt}, for conducting online test-time adaptation of semantic segmentation models. In TransAdapt, we first pretrain the semantic segmentation model with both supervised and unsupervised loss functions,
where the final supervised  segmentation predictions are mapped from the unsupervised predictions. Specifically, given a segmentation prediction head that is trained with an unsupervised loss, we leverage a transformer module to convert the unsupervised predictions to the final predictions via a linear mapping. During test, only the unsupervised head receives training signals to update the model without incurring costly updates on the transformer. The transformer module leverages global context within features to generate the unsupervised-to-supervised mapping and also empirically produces better recognition performance compared to non-transformer variants.

During online test-time adaptation, we propose to use transformation consistency (TC) as the unsupervised loss for updating the model. By utilizing TC, we avoid relying on noisy and inaccurate pseudo-labels of target domain images.  Specifically, we consider two types of transformation: (a) photometric transformations which enforce model invariance to non-essential visual appearance changes and (b) geometric transformations which enforce the model's prediction to be equivariant to the input, w.r.t. certain geometric transformations. Using our transformer based pre-training and/or transformation consistency based adaptation, we produced improved segmentation on the three cross-dataset benchmarks. Our main contributions are summarized as follows: 
\begin{itemize}
\item We propose a plug-and-play transformer module on top of a segmentation network that produces supervised segmentation outputs from unsupervised ones, thus allowing generalization without requiring adaptation. 
\item We devise a test-time adaptation loss using photometric and geometric transformation consistency.
\item Finally, we evaluate the effectiveness of our framework against existing but compatible unsupervised and adaptive baselines and establish three cross-dataset benchmarks to facilitate future research.
\end{itemize}

\section{Proposed Framework}
\subsection{Task Description}
We consider the availability of labeled source domain data $\mathcal{X}^{src} = \{ (x^{src}_i, y^{src}_i) \}_{i=1}^{N_{src}}$ with input images $x^{src}_i \in \mathbb{R}^{H\times W\times 3}$ and their segmentation maps $y^{src}_i \in \mathbb{R}^{H\times W \times L}$. Here, $H$ and $W$ are the height and width of the input image and $L$ is the number of class labels. This labeled source domain data is used to pre-train a model. For online test-time adaptation, we need to adapt the model to a sequence of unlabeled images $\mathcal{X}^{tgt} = \{ ( x^{tgt}_i )\}_{i=1}^{N_{tgt}}$ from the target domain with the same set of classes as the source domain. It is to be noted that the sequence of images are not necessarily adjacent frames of a video.
{Also, the model can be decomposed into a feature extractor $F$, a prediction head $H_p$ and optionally our proposed transformer module $T_f$.}

\subsection{Pre-training with Transformer Module}
{We use the transformer module to find the relationship between supervised and unsupervised semantic segmentation maps as described in Fig.~\ref{fig:pretrain}.}
Specifically, consider an input image $x_i \in \mathbb{R}^{H\times W\times 3}$, which produces a feature map $f_i \in \mathbb{R}^{H' \times W' \times C}$ such that $f_i = F(x_i)$. This feature map when passed through a prediction head, produces output logits $o_i \in \mathbb{R}^{H\times W\times L}$ such that $o_i = H_p(f_i)$. A softmax operation is applied on these logits to obtain a segmentation probability map $p_i \in \mathbb{R}^{H\times W\times L}$. For end-to-end training of $F$ and $H_p$, we can use cross entropy loss $\mathcal{L}_{XEnt}(p_i,y_i)$ between the predicted probability maps $p_i$ and ground truth segmentation labels $y_i$.

In our proposed framework, we aim to learn the relationship between supervised and unsupervised predictions which would facilitate test-time adaptation from unlabeled image sequences.
{Here, we use the output from the prediction head as unsupervised logit $o_i^{u}$.}
The feature map $f_i$ is then used as conditioning input for a transformer decoder module to construct the keys and the values. The transformer decoder uses learnable queries $q^{\text{s}}\in \mathbb{R}^{L\times C}$ which are $C$ dimensional vector representations of $L$ categories to be identified for the supervised output.
To generate keys and values, the patches are obtained from the feature map $f_i$ which are then flattened to produce $n$ tokens $t_i \in \mathbb{R}^{n \times C}$. 
These tokens are then fed into the multi-head attention stage of the transformer decoder followed by a feed forward network. To understand the multi-head attention scheme, we first mention the single-head attention mechanism which is as follows:
$
  q = q^{\text{s}}W^{q},\:k_i = t_i W^{k},\:v_i = t_i W^{v}
$
 where $W^{q}, W^{k}, W^{v} \in \mathbb{R}^{C\times C}$ are weight matrices to produce linear representations of the raw tokens. These processed tokens are then used to produce the attention operation $\text{Att}(\cdot)$ such that
$
     \text{Att}(q,t_i,t_i) = \text{Softmax}(qk^{T}_i)v_i
$.
For the multi-head attention operation $\text{MHAtt}(\cdot)$ having $M$ heads, $q^{\text{s}}$ and $t_i$ are split into $M$ parts $q^{\text{s}}_1,\dots,q^{\text{s}}_M$ and $t_{i,1},\dots, t_{i,M}$, where dimension of each split is $C' = C/M$. Attention operation is applied over all such splits to produce 
\begin{equation}\label{eq:split}
  \tilde{q}^{\text{s}}_i = \left[\text{Att}_1(q^{\text{s}}_1, t_{i,1}, t_{i,1});\dots;\text{Att}_M(q^{\text{s}}_M, t_{i,M}, t_{i,M})\right]
\end{equation}
\begin{equation}\label{eq:mha}
  \text{MHAtt}\left(q,t_i,t_i \right) = \mathrm{LN}\left(q^{\text{s}} + \mathrm{DO}\left(\tilde{q}^{\text{s}}_iW\right)\right)
\end{equation}
where $\mathrm{LN}$ is layer normalization~\cite{ba2016layer}, $\mathrm{DO}$ is the dropout operation~\cite{JMLR:v15:srivastava14a} and $W \in \mathbb{R}^{C\times C}$ is a linear mapping. The output of the multi-head attention mechanism is passed through a two layer feed-forward network where each layer consists of a linear mapping followed by dropout, residual connection and layer normalization similar to Eq.~\ref{eq:mha}.
Alternating multi-head attention and feed-forward networks can produce multiple layers of the transformer decoder. The output of the transformer decoder will produce a representation $q^{\text{o}}_i \in \mathbb{R}^{L\times C}$ for the transformer decoder input $q^{\text{s}}$. $q^{\text{o}}_i$ thus consists of $C$-dimensional vector representation of each of the $L$ classes conditioned on the feature map of the input image. This representation needs to be mapped to a $L$-dimensional space for it to produce a $L \times L$ weight matrix that relates supervised and unsupervised logits. Hence, we apply the following operations
\begin{equation}\label{eq:out_sup}
W^{s}_{u}=\mathrm{Softmax}\left(q^{\text{o}}_i U\right) \quad,\quad o_i^{s}=o_i^{u} W^{sT}_{u}.
\end{equation}
{Here, $U \in \mathbb{R}^{C\times L}$ is a projection layer, $W^{s}_{u}$ is the transfer matrix and $o_i^{s}$ are the supervised output logits. Softmax is then applied to these logits to obtain the segmentation probability map $p_i^{s} \in \mathbb{R}^{H\times W\times L}$.} For end-to-end training of the whole model, we can use cross entropy loss $\mathcal{L}_{XEnt}(p_i^{s},y_i)$ between the predicted probability maps $p_i^{s}$ and ground truth segmentation labels $y_i$. For training the model using unsupervised logits $o_i^{u}$, we can use an unsupervised loss $\mathcal{L}_{USup}(o_i^{u})$. This unsupervised loss can possibly be one of the losses used for test-time adaptation such as min-entropy~\cite{WangICLR21TENTFullyTestTimeAdaptEntropyMin}, max-squares~\cite{ChenICCV19DASemSegmentationMaximumSquaresLoss} or our proposed transformation consistency loss. We can thus train the whole network using the total loss
\begin{equation}\label{eq:trainloss}
    \mathcal{L}_{Tot}(x_i, y_i) = \mathcal{L}_{XEnt}(p_i^{s},y_i) + \lambda \mathcal{L}_{USup}(o_i^{u})
\end{equation}
By minimizing this training loss with the source domain data, we can learn the mapping between unsupervised and supervised segmentation predictions. During online test-time adaptation over a sample $x$, the transformer module is kept frozen and we use the output $o^{u}$ of the unsupervised head for obtaining the unsupervised loss $\mathcal{L}_{TTA}(o^{u})$ to be used for updating the model parameters. After adaptation is complete, we use the supervised head outputs $o^{s}$ for evaluation purposes.
{In the next sub-section, we explain our proposed transformation consistency loss as an unsupervised loss for test-time adaptation.}


\subsection{Adaptation with Transformation Consistency}
{To resolve erroneous updates due to noisy pseudo-labels of single images, we apply transformation consistency as a loss for online test-time adaptation using invariance and equivariance property of different transformation types.} {We use two transformation types - photometric (grayscale, jitter, blur) and geometric (cropping, rotations, shuffling), for invariance and equivariance respectively.}


Specifically, let's consider a test image $x$ and a sampled photometric transformation $\mathcal{A}_p(\cdot)$. When this transformation is applied on a test image, it will produce a transformed image $\tilde{x} = \mathcal{A}_p(x)$. For both the original input image $x$ and its transformation $\tilde{x}$, we produce unsupervised output logits $o^{u} = H_p(F(x))$ and $\tilde{o}^{u} = H_p(F(\tilde{x}))$ respectively. To minimize the difference between $o^{u}$ and $\tilde{o}^{u}$, we can use discrepancy loss term $\mathcal{L}_p(o^{u},\tilde{o}^{u})$. Possible discrepancies are $L1$ or $L2$ distances.
Specifically, let's also consider a sampled geometric transformation $\mathcal{A}_g(\cdot)$. When the geometric transformation is applied on the test image, it will produce a transformed image $\hat{x} = \mathcal{A}_g(x)$. For both the original input image $x$ and its transformation $\hat{x}$, we produce unsupervised output logits $o^{u} = H_p(F(x))$ and $\hat{o}^{u} = H_p(F(\hat{x}))$ respectively. To enforce equivariance, we minimize the difference between $\hat{o}^{u}$ and the transformed logits $\mathcal{A}_g({o}^{u})$ by using a discrepancy loss term $\mathcal{L}_g(\mathcal{A}_g({o^{u}}),\hat{o}^{u})$. The discrepancy will be the same as used for photometric transformation consistency loss. For adapting the model on the test sample $x$, we use both photometric and geometric transformation consistency losses as follows
\begin{equation}\label{eq:consistency}
\mathcal{L}_{TTA}(x) = \mathcal{L}_p(o^{u},\tilde{o}^{u}) + \mathcal{L}_g(\mathcal{A}_g(o^{u}),\hat{o}^{u})
\end{equation}
Once the model is adapted using $\mathcal{L}_{TTA}(x)$, we infer the output predictions with the supervised head using Eq.~\ref{eq:out_sup}. We reiterate that during test-time adaptation, the back-propagated gradients through the unsupervised head do not affect the transformer module and hence it remains frozen throughout. When the transformer module is not used, the model has a single head. $o^{u}$ and subsequently $\mathcal{L}_{TTA}(x)$ is processed through the single head for adaptation, and inference is carried out through that single head only. We summarize our pre-training and adaptation step in Algorithm 1.

\setlength{\textfloatsep}{10pt} 
\begin{algorithm}[h]
\SetAlgoLined
 \textbf{Given:} Source dataset $\mathcal{X}^{src} = \{ (x^{src}_i, y^{src}_i) \}_{i=1}^{N_{src}}$ \& Target dataset sequence $\mathcal{X}^{tgt} = \{ ( x^{tgt}_i )\}_{i=1}^{N_{tgt}}$   \\
  \textbf{Step 1:} Pre-train model on $\mathcal{X}^{src}$ \\
 \textbf{For} each sample $(x^{src}_i, y^{src}_i)$ from sampled batch of $\mathcal{X}^{src}$ \\
 \quad Gradient update of Eq.~\ref{eq:trainloss} w.r.t. $F$, $H_p$, $T_f$, $U$, $q^s$, $W^{q,v,k}$ \\
 \textbf{Step 2:} Adaptation and Evaluation on $\mathcal{X}^{tgt}$ \\
 \textbf{For} each sample $x^{tgt}_i$ from $\mathcal{X}^{tgt}$\\
 \quad Gradient update of Eq.~\ref{eq:consistency} w.r.t $F$ and $H_p$\\
 \quad Predict segmentation map of $x^{tgt}_i$ using Eq.~\ref{eq:out_sup} \\
\caption{TransAdapt framework}
\end{algorithm}

\section{Experimental Results} 

\subsection{Experiment Details}
We evaluate our framework using three cross-dataset settings as in~\cite{ChenICCV19DASemSegmentationMaximumSquaresLoss}: GTA5 (Synthetic)~\cite{RichterECCV16PlayingForData} $\rightarrow$ Cityscapes (Real)~\cite{CordtsCVPR16CityscapesDataset}, SYNTHIA (Synthetic)~\cite{RosCVPR16SYNTHIADataset} $\rightarrow$ Cityscapes and Cityscapes $\rightarrow$ Cross-City (Real)~\cite{ChenICCV17NoMoreDiscriminationCrossCityAdaptationRoadSceneSegmenters} using online adaptation on test-set instead of offline adaptation. For all evaluation metrics, we use mean Intersection-over-Union (mIoU).
For the segmentation model, we use DeepLab-V2 ResNet-101~\cite{ChenICCV19DASemSegmentationMaximumSquaresLoss} trained on each of the source datasets. For our proposed transformer module, we use a 1-layer decoder without positional encoding, and output of block 3 of ResNet is used for the input of of the transformer module. This transformer module is trained with the segmentation network together by the loss defined in Eq.~\ref{eq:trainloss}.
Unless explicitly mentioned, we set $\lambda=0.1$ and max squares~\cite{ChenICCV19DASemSegmentationMaximumSquaresLoss} as the unsupervised loss. For the transformation consistency loss in Eq.~\ref{eq:consistency}, $L2$ distance is used as the default metric for both $\mathcal{L}_p(\cdot)$ and $\mathcal{L}_g(\cdot)$. For the adaptation, we use SGD with learning rate of $1e-4$ and update only batch-norm parameters for only1 iteration per sample as more updates cause performance degradation.

\subsection{Comparison Studies}
We compare against some popular domain adaptive methods~\cite{WangICLR21TENTFullyTestTimeAdaptEntropyMin, ChenICCV19DASemSegmentationMaximumSquaresLoss, wang2021give,prabhu2021s4t} and unsupervised segmentation methods~\cite{kanezaki2018unsupervised,kim2020unsupervised}. We also compare against recently proposed online adaptive methods: AuxAdapt~\cite{auxadapt} , Batch Norm update~\cite{schneider2020improving}, Style Transfer variants~\cite{yoo2019photorealistic}, of which the latter two have been used in the OASIS~\cite{Volpi_2022_CVPR} benchmark. 
{We further evaluate against two proposed baselines:
(a) Selective Cross Entropy: we apply cross-entropy loss on only those pixels whose confidence score is greater than 0.8.
(b) Special Cross Entropy: We use the original test image and its photometric transformed image and apply weighted cross entropy loss where the weight depends on agreement of predictions from both the images.
}

\begin{figure}[h]
\centering
\includegraphics[width=0.99\linewidth]{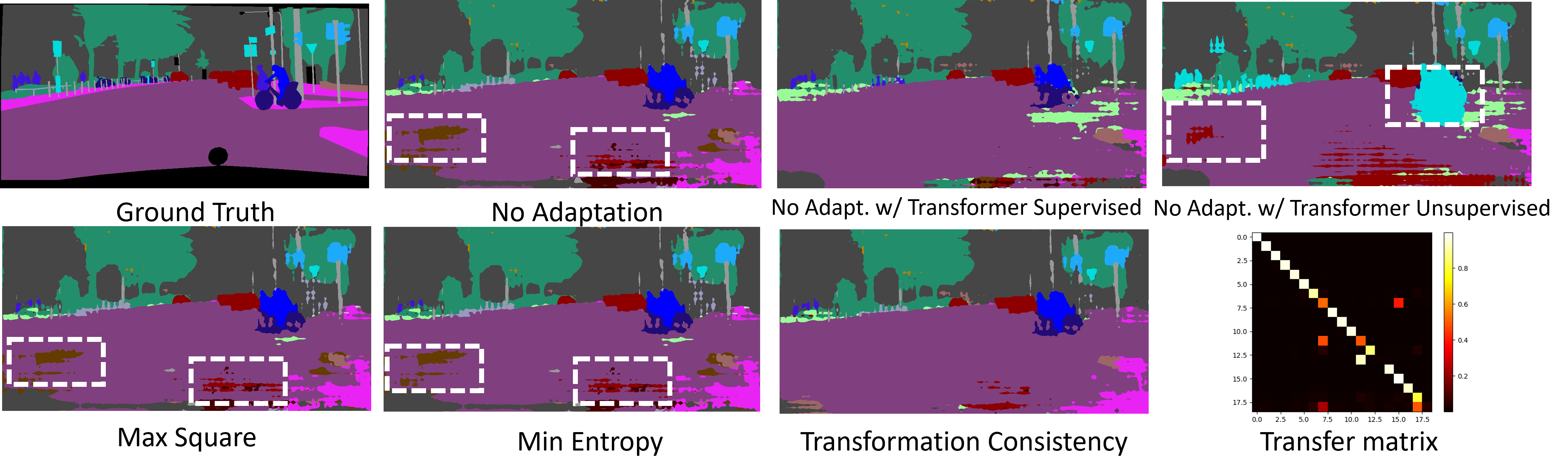}
\vspace{-8pt}
\caption{\small Segmentation masks for different methods and transfer matrix for our proposed method. White borders show that our consistency and transformer-based approach produces better segmentation.}
\label{fig:segmap}
\end{figure}

Table~\ref{tab_003} shows that for all cross-dataset setups, our proposed transformation consistency method achieves mostly better performance compared to other methods.
{Interestingly, using the transformer block, we observe mIoU improvement even without adaptation.  When we apply our consistency method to the model with the transformer block, there is further mIoU improvement. } Figure \ref{fig:segmap} illustrates visualization of predicted segmentation masks. Our proposed transformer module improves performance in case of no adaptation and our proposed transformation consistency method is better than other adaptation techniques. In Fig.~\ref{fig:segmap}, we highlight that the unsupervised segmentation map produces more errors compared to the supervised segmentation map. Furthermore, presence of non-zero values on off-diagonal elements of the transfer matrix $W^{s}_{u}$ suggests that supervised prediction is related through a combination of different unsupervised predicted categories.

\begin{table}[t]
	\centering
		\caption{\small Results for GTA5 (GTA)-to-Cityscapes (CS), SYNTHIA (SYN)-to-Cityscapes (CS) and Cityscapes (CS)-to-CrossCity (Rome, Rio, Tokyo, Taipei) experiments.}\vspace{-8pt}
	\resizebox{0.48\textwidth}{!}{    
{
			\begin{tabular}{l| c | c c c c c c }
				\toprule
				\multicolumn{8}{c}{Source Dataset $\to$ Target Dataset} \\
				\midrule
				Method & Backbone &
				\rotatebox{90}{GTA $\to$ CS} & \rotatebox{90}{SYN $\to$ CS} &
				\rotatebox{90}{CS $\to$ Rome} & \rotatebox{90}{CS $\to$ Rio} & \rotatebox{90}{CS $\to$ Tokyo} & \rotatebox{90}{CS $\to$ Taipei}  \\
				\hline
				No Adaptation
				&  &33.68 &28.66  &50.19 & 48.91 & 47.78    & 45.30    \\
				Min Entropy~\cite{WangICLR21TENTFullyTestTimeAdaptEntropyMin}
				&  &36.20 &31.77  &50.74      & 49.79      & 48.11         & 45.59         \\
				Max Squares~\cite{ChenICCV19DASemSegmentationMaximumSquaresLoss}
				&  &37.24 &31.38  &50.73      & 49.72      & 48.00         & 45.56         \\
				Focal Entropy~\cite{wang2021give}
				&  &33.78 &28.85  &50.6       & 49.48      & 47.90         & 45.50         \\
				Sel. Cross Entropy
				&  &37.28 &30.60  &50.62      & 49.49      & 47.91         & 45.50          \\
				S4T~\cite{prabhu2021s4t}        
				& DeepLab-V2 &35.66 &30.89  &50.51      & 49.43      & 47.87         & 45.47         \\
				Spec Cross Entropy
				& RN-101 &36.22 &28.97  &50.63      & 49.49      & 47.91         & 45.50          \\
				Super Pixel~\cite{kanezaki2018unsupervised}
				&  &36.31 &31.87  &50.60      & 49.56      & 47.91         & 45.52        \\
				Spatial Cont.~\cite{kim2020unsupervised}    
				&  &36.81 &32.53  &50.55      & 49.40      & 47.88         & 45.65         \\
				AuxAdapt~\cite{auxadapt}  
				&  &36.63 &28.74  & 50.30 & 49.74 & 48.13 & \textbf{46.38} \\
                Style Transfer Rand~\cite{yoo2019photorealistic,Volpi_2022_CVPR}  
                &  &35.74 &33.34  & 47.20 & 46.49 & 45.54 & 43.13 \\
                Batch Norm~\cite{schneider2020improving}  
                &  &33.68 &28.66  & 50.20 & 48.92 & 47.78 & 45.30 \\
				Trans. Cons. (Ours)
				&  &\textbf{37.83} &\textbf{33.72}  & \textbf{50.82}      & \textbf{50.43}      & \textbf{48.20}         & {45.88}         \\
				\hline
				No Adaptation
				& &35.61 &31.25  & 51.14      & 49.01      & 47.48         & 47.45          \\
				Min Entropy~\cite{WangICLR21TENTFullyTestTimeAdaptEntropyMin}
                &  DeepLab-V2 &36.82 &31.53  & 51.61      & 49.35      & 47.64         & 47.58         \\
                Max Squares~\cite{ChenICCV19DASemSegmentationMaximumSquaresLoss}
                &  RN-101 &36.59 &31.67  & 51.53      & 49.31      & 47.61         & 47.58         \\ Trans. Cons. (Ours)
                & + Transformer &\textbf{37.08} &\textbf{33.02}  & \textbf{51.67}      & \textbf{49.66}      & \textbf{47.91}         & \textbf{47.72}         \\
				\bottomrule    
	\end{tabular}}}

	\label{tab_003}
\end{table}

\subsection{Ablation Studies and Analyses}

\begin{figure}[h]
\centering
\includegraphics[width=0.99\linewidth]{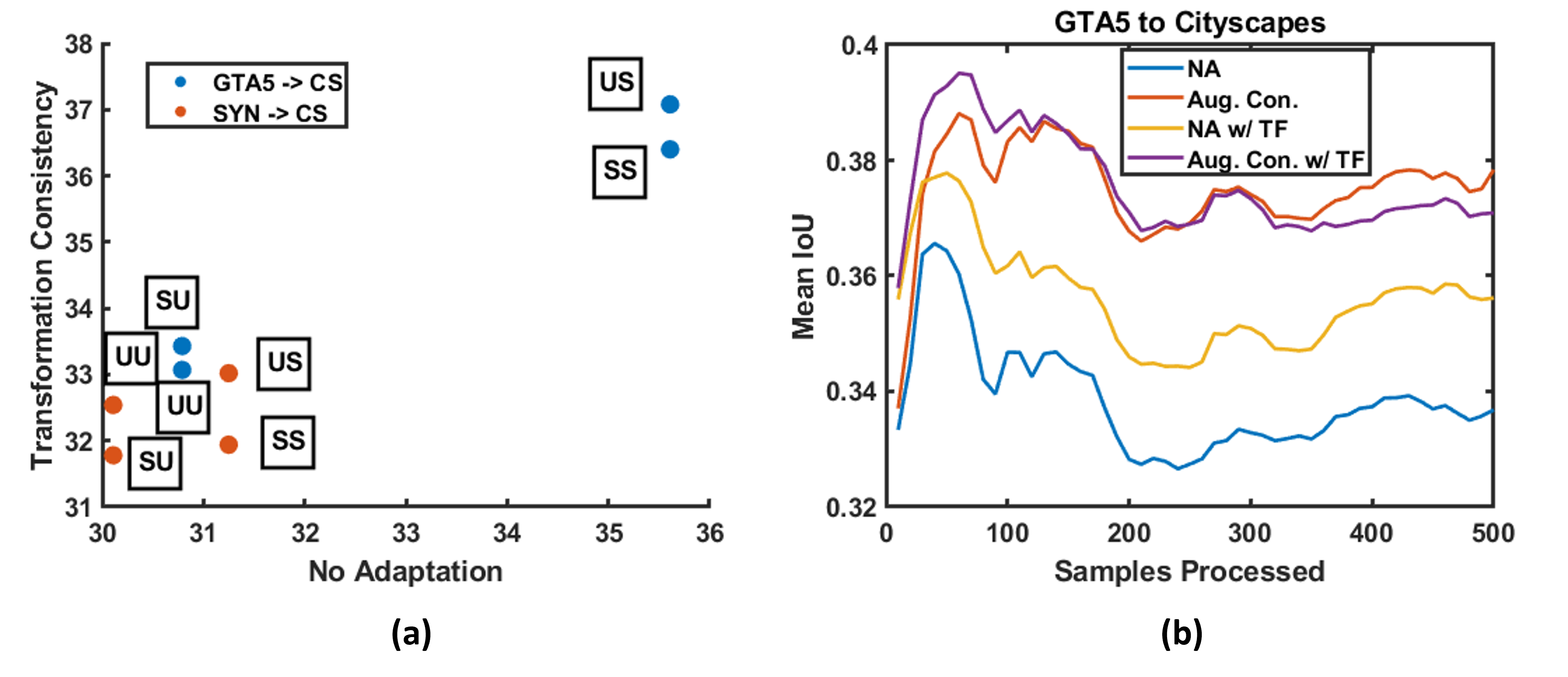}\\
\vspace{-8pt}
\caption{\small (a) Effect on mIoU by varying update and inference heads. XY implies update using X head and inference using Y head. (b) Performance evolution as sample sequence is processed by the model. }
\label{fig:iou_evol}
\end{figure}

In Fig.~\ref{fig:iou_evol} (a), we report results when alternative ways for update and inference is done in test-time adaptation. Here, XY denotes that X is the update pass and Y is the inference pass. For example, the default setting of US is that update is done using unsupervised (U) head and the supervised (S) head is used for predicting results. Results show that this default setting is optimal and it surpasses over other configurations.
Fig.~\ref{fig:iou_evol} (b) shows that the proposed method achieves an overall performance improvement. As observed by the blue curve (No Adaptation), mIoU increases slightly until 80 samples and then reduces. This implies that performance with no adaptation varies across a sequence of samples.
However, the curves of our proposed methods exhibit better performance than the No Adaptation curve for all sequences.
We observe significant gains using both the transformation consistency loss and the transformer block.



To analyse which feature layer is useful for input to our proposed transformer decoder module, we consider different layer outputs (features) as input to the transformer module. Table \ref{tab:feats_trans} illustrates performance based on different features. `Out' represents using logits layer {from prediction head output} and `Orig' represents the use of default feature map (described in Sec 3.1). 
{`F+1' uses output from the next layer (Block 4) of the `Orig' setting. Likewise, `F-1' \& `F-2' uses output of one and two layers before 'Orig' setting, which are outputs of Block 2 and Block 1, respectively.}
Results suggest that `Orig' produces the best result in case of GTA5 $\to$ Cityscapes, but `F-1' produces the best performance in case of SYNTHIA $\to$ Cityscapes. Furthermore, all variations perform better than not having transformer (No TF), a configuration where we have independent supervised and unsupervised heads. {We also analysed other design choices such as use of reconstruction loss during training (Recon), use of convolutional block (Conv) or additional linear layer (Conv+Lin) for generating transfer matrix. All these variations still perform poor compared to our proposed transformer variants.}

\begin{table}[t]
	\centering
		\caption{\small {{Effect of different design choices and varying feature map inputs to our transformer module with \textit{No adaptation}}. 
		{Here, G, C, and S denote GTA5, Cityscapes, and SYNTHIA dataset respectively.}
		}\vspace{-8pt}
}
	\resizebox{0.47\textwidth}{!}{    
{
			\begin{tabular}{l| c  c c c c  c c c c }
				\midrule
				Setup & Orig & \rotatebox{360}{F+1} & \rotatebox{360}{Out} & \rotatebox{360}{F-1} & \rotatebox{360}{F-2}  & No TF & Recon & Conv & Conv+Lin\\
				\hline
				G$\to$C
				& 35.61 & 34.36 & 31.18 & 33.55 & 34.41 & 29.28 &31.22 &26.07 &26.08  \\
			    S$\to$C
				& 31.25 & 30.86     & 30.48     & 32.01        & 32.45       & 29.68 &30.87 &26.32 &27.12  \\
				\bottomrule    
	\end{tabular}}}

	\label{tab:feats_trans}
\end{table}

We also analysed the effect of varying the  unsupervised loss function $\mathcal{L}_{USup}$ in Eq.~\eqref{eq:trainloss} applied during pre-training phase of transformer block. We perform experiments with different objectives: transformation consistency, max squares and min entropy as they represent competitive and common methods reported in previous experiments. From Table~\ref{tab:incr_aug} last columns, we notice that transformation consistency performs the best in case of GTA5 $\to$ Cityscapes and max squares performs the best in case of SYNTHIA $\to$ Cityscapes. We also reported in Table~\ref{tab:incr_aug}, the results of increasing transformed samples during adaptation for transformation consistency, max squares and min entropy. We observe that as number of transformations increases, performance improves in case of transformation consistency, especially up to 4 but saturates after that. However, increasing number of transformed samples do not increase the performance for min entropy and max squares based methods.



\begin{table}[t]
	\centering
		\caption{\small Effect of increasing number of transformations for GTA5$\to$Cityscapes (left) and SYNTHIA$\to$Cityscapes (right) respectively. The last column (Loss) considers no adaptation baselines when using corresponding unsupervised loss during pretraining}\vspace{-8pt}
	\resizebox{0.5\textwidth}{!}{    
{
			\begin{tabular}{l| c  c c c c|  c  c c c c}
				\midrule
				Setup & 1 & \rotatebox{360}{2} & \rotatebox{360}{4} & \rotatebox{360}{8} & Loss & 1 & \rotatebox{360}{2} & \rotatebox{360}{4} & \rotatebox{360}{8} & Loss \\
				\hline
				Min Entropy
				& 36.74 & 36.29 & 36.14 & 36.23 & 35.61 	& 30.17 & 30.80 & 30.78 & 30.69 & 31.25  \\
			    Max Squares
				& 36.97 & 36.65     & 36.51    & 36.34   & 34.18 	& 30.89 & 30.28     &  30.54    & 30.61    & 30.31     \\
				Trans. Cons.
				& 37.83 & 38.39     & 38.80    & 38.61   & 36.70 	& 33.72 & 33.73    & 34.08    & 33.98  & 30.18   \\
				\bottomrule    
	\end{tabular}}}

	\label{tab:incr_aug}
\end{table}

\section{Conclusion}
In this paper, we propose a framework for online test-time adaptive semantic segmentation. Our method consists of learning a transformer module to map unsupervised predictions to supervised predictions. We also proposed transformation consistency as a fine-tuning objective to adapt our model on online unlabeled target domain data. Experimental studies showed that our proposed framework outperforms other competitive methods both quantitatively and qualitatively. Furthermore, we carried out extensive analyses to find out design choices and ablations of our framework that affect segmentation performance.

\vfill\pagebreak

\clearpage

\bibliographystyle{IEEEbib}
\bibliography{egbib_short}

\clearpage

\appendix

\section{Additional Implementation Details}
Here, we describe some of the experimental details that have been omitted in the original paper due to space limits. The pre-training scheme used is the same as ~\cite{ChenICCV19DASemSegmentationMaximumSquaresLoss}. We used a batch size of 4 and the model is trained on single V100 GPUs. The polynomial learning rate scheduler was used with a power of 0.9. For the transformer decoder module, we use 4 heads.

For the the color jitter operation in computing the transformation consistency loss, we used a value of 0.75 for all of brightness, contrast, saturation and hue. For the random crop operation, we used crop ratio of 0.5. For the pixel shuffle operation, we use patches of size 256 that are randomly shuffled. Also, we found that updating layers beyond batch norm and updating for more than 1 iteration per sample yielded poorer results and hence we don't report them.

We also proposed two adaptation losses: Selective Cross Entropy and Special Cross Entropy.
The Selective Cross Entropy loss for a pixel is defined as follows:
\begin{equation}\label{eq:selxent}
\mathcal{L}_{sel} = -\sum_{l}^{L} I(l,p_l)\log{p_l}
\end{equation}
where $p_l$ is the probability of class $l$ among $L$ classes. $I(l,p_l)$ is an indicator function which is 1 if and only if $p_l$ is greater than $0.8$ and also if $l$ is the pseudo-label for the pixel. Otherwise, $I(l,p_l)$ is 0. The pseudo-label is found by finding the class $l$ which maximizes $p_l$ over all the possible $L$ classes. The total loss for adaptation is found by averaging $\mathcal{L}_{sel}$ over all the pixels.

The Special Cross Entropy loss for a pixel is defined using a transformation consistency scheme. Consider an input image $x$ and it's photometric transformation $\tilde{x}$. For a particular pixel, let $x$ and $\tilde{x}$ generate probabilities $p_l$ and $\tilde{p_l}$ for a particular pixel and class $l$. Then, the loss is defined as
\begin{equation}\label{eq:selxent}
\mathcal{L}_{spc} = -\sum_{l}^{L} w(p_l, \tilde{p_l})\log{\tilde{p_l}}.
\end{equation}
Here, $w(p_l, \tilde{p_l})$ is 1 if pseudo-labels extracted from $p_l$ and $\tilde{p_l}$ by argmax operation are the same. Otherwise, we use the following: $w(p_l, \tilde{p_l}) = \exp{(-||p_l - \tilde{p_l}||_2^{2})}$. The total loss for adaptation is found by averaging $\mathcal{L}_{spc}$ over all the pixels. 

We also experimented with other architecture designs, the results of which are shown in Table 4 of the main paper. The first architecture modification uses reconstruction (Recon), where the input images are reconstructed using the $L2$ loss. The reconstruction network is applied on top of the default feature map and consists of two modules. The first module is a duplicate of the fourth block of the ResNet architecture. The second module is a duplicate of the classifier module except that it outputs three channels for reconstructed input images. Furthermore, we also explored non transformer-based architectures for learning the mapping from unsupervised to supervised predictions. The first type (Conv) consists of two convolutional blocks where each block consists of a convolution layer and batch normalization connected through a ReLU non-linearity. Spatial average pooling is applied and reshaped to the transfer matrix dimension. In this case, feature channel size changes from 1024 to 512 to 361 which is then reshaped to a 19 times 19 transfer matrix, when the number of classes is 19. The second type (Conv + Lin) of mapping architecture also consists of two convolutional blocks. However, there is an additional linear layer after average pooling. In this case, feature channel size changes from 1024 to 512 to 256 which is then mapped to a dimension of 361 by a linear layer. That feature is then reshaped to a 19 times 19 transfer matrix. For tasks containing 13 classes, the transfer matrix size will be 13 times 13.

\section{Additional Experiments}
In this section, we report results of additional experiments. For pre-training the transformer module, when we use positional encoding, it produced a relatively poorer no adaptation recognition performance of 34.54\% mIOU compared to the default of 35.61\% mIOU for GTA5 $\to$ Cityscapes. For SYNTHIA $\to$ Cityscapes, using positional encoding produced slightly better recognition performance of 31.77\% mIOU compared to the default of 31.25\% mIOU.

In Table~\ref{tab:loss}, we report results of the variation in the distance metric used for $\mathcal{L}_p(\cdot)$ and $\mathcal{L}_g(\cdot)$ in Eq.5 of the main paper. The default metric is $L2$ distance over the logits $o$. Another variation includes $L1$ loss over the logits. Alternatively, $L1$ or $L2$ distances can be applied on the probabilities that are obtained by applying softmax on the logits. Results on both benchmarks GTA5$\to$Cityscapes and SYNTHIA$\to$Cityscapes show that the default metric is the most optimal. Applying $L1$/$L2$ distances over probabilities show poorer results possibly because probabilities have lower ranges and do not provide higher gradient magnitudes for model update. We also tried applying KL divergence loss between the probabilities but it yielded very poor mIoU i.e. 1.75 \%  and 2.08 \% for the GTA5$\to$Cityscapes and SYNTHIA$\to$Cityscapes setups respectively.

\begin{table}[h]
	\centering
		\caption{Effect of using different consistency losses during test-time adaptation.}
	\resizebox{0.45\textwidth}{!}{    
{
			\begin{tabular}{c| c  c c c }
				\midrule
				Setup & L2 Log. & L1 Log. & L2 Prob. & L1 Prob. \\
				\hline
				GTA5$\to$Cityscapes
				& 37.83 & 36.62 & 34.45 & 34.59  \\
			    SYNTHIA$\to$Cityscapes
				& 33.72 & 32.65     & 30.98     & 31.06  \\
				\bottomrule    
	\end{tabular}}}

	\label{tab:loss}
\end{table}

In Table~\ref{tab:lamda}, we vary $\lambda$, which is the weight on the unsupervised loss $L_{USup}$ used for pre-training and defined in Eq.4 of the main paper. This table shows no adaptation results using our transformer module when evaluated using GTA5$\to$Cityscapes and SYNTHIA$\to$Cityscapes benchmarks. We obtain optimal results for the default value of $\lambda=0.1$. Higher values of $\lambda$ causes larger drop in performance probably because the network tries to focus less on learning using supervisory signals compared to learning from self-supervision.

\begin{table}[h]
	\centering
		\caption{Effect of using different $\lambda$ without adaptation.}
	\resizebox{0.45\textwidth}{!}{    
{
			\begin{tabular}{c | c  c c c }
				\midrule
				$\lambda$ & 0.01 & 0.1 & 1 & 10 \\
				\hline
				GTA5$\to$Cityscapes
				& 35.22 & 35.61 & 33.39 & 33.19  \\
			    SYNTHIA$\to$Cityscapes
				& 30.90 & 31.25     & 30.47     & 30.01  \\
				\bottomrule    
	\end{tabular}}}

	\label{tab:lamda}
\end{table}

In Table~\ref{tab:layer}, we study the effect of increasing the size of the transformer decoder module, when evaluating using GTA5$\to$Cityscapes and SYNTHIA$\to$Cityscapes benchmarks without adaptation. Results show that having 1 layer for the transformer decoder produces the optimal performance. Surprisingly, using 4 layers for the transformer module produces large drop in performance (14.37) for GTA5$\to$Cityscapes. This maybe due to severe overfitting of the transformer module, when it is pre-trained on the GTA5 dataset.

\begin{table}[h]
	\centering
		\caption{Effect of using different number of layers for the transformer decoder without adaptation.}
	\resizebox{0.45\textwidth}{!}{    
{
			\begin{tabular}{c | c  c c c }
				\midrule
				No. of Layers & 1 & 2 & 3 & 4 \\
				\hline
				GTA5$\to$Cityscapes
				& 35.61 & 33.49 & 34.44 & 14.37  \\
			    SYNTHIA$\to$Cityscapes
				& 31.25 & 30.71    & 31.18     & 31.20  \\
				\bottomrule    
	\end{tabular}}}

	\label{tab:layer}
\end{table}

In Table~\ref{tab:std}, we report accuracy results of methods that use transformation during test-time adaptation. The accuracy numbers are replicated from Table 1, 2 and 3 in the main paper except that we also report standard deviation across 5 runs. This is important to show because the transformations are generated randomly and they produce different results across different runs. From the results in Table~\ref{tab:std}, we observe that all the methods are stable across different runs with low standard deviations. Our proposed method has comparatively higher standard deviation but the improvement is still statistically significant compared to other methods.

\begin{table}[h]
	\centering
		\caption{\small Segmentation mIOU results for transformation-based approaches along with standard deviation. Here CS stands for Cityscapes and SYN stands for SYNTHIA.}\vspace{-8pt}
	\resizebox{0.45\textwidth}{!}{    
{
			\begin{tabular}{l| c | c c c c c c}
				\toprule
				\multicolumn{8}{c}{Source$\to$Target} \\
				\midrule
				Method & Backbone & \rotatebox{90}{CS$\to$Rome} & \rotatebox{90}{CS$\to$Rio} & \rotatebox{90}{CS$\to$Tokyo} & \rotatebox{90}{CS$\to$Taipei} & \rotatebox{90}{GTA$\to$CS}
				& \rotatebox{90}{SYN$\to$CS}  \\
				\hline
				S4T~\cite{prabhu2021s4t}    
				& DeepLab-V2 & \makecell{50.51\\(0.02)}      & \makecell{49.43\\(0.05)}      & \makecell{47.87\\(0.01)}        & \makecell{45.47\\(0.02)} & \makecell{35.66\\(0.03)} & \makecell{30.89\\(0.05)}        \\
				Spec Cross Entropy
				& RN-101 & \makecell{50.63\\(0.02)}      & \makecell{49.49\\(0.01)}      & \makecell{47.91\\(0.01)}         & \makecell{45.50\\(0.01)}  & \makecell{36.22\\(0.03)} & \makecell{28.97\\(0.01)}         \\
				Trans. Cons. (Ours)
				&  & \makecell{{50.82}\\(0.01)}      & \makecell{{50.43}\\(0.12)}      & \makecell{{48.20}\\(0.09)}         & \makecell{{45.88}\\(0.08)}  & \makecell{37.83\\(0.12)} & \makecell{33.72\\(0.09)}       \\
				\hline
             Trans. Cons. (Ours)
                & + Transformer & \makecell{{51.67}\\(0.01)}      & \makecell{{49.66}\\(0.08)}      & \makecell{{47.91}\\(0.07)}         & \makecell{{47.72}\\(0.08)}  & \makecell{37.08\\(0.09)} & \makecell{33.02\\(0.06)}       \\
				\bottomrule    
	\end{tabular}}}

	\label{tab:std}
\end{table}



Quantitative results on the individual effects of photometric and geometric transformation consistency losses are shown in Table~\ref{tab:photogeo}. Results show that removing either of the photometric transformation consistency loss term or the geometric transformation consistency loss term produces drop in recognition performance across all the datasets. This shows that a combination of both these consistencies are required for improved performance.

\begin{table}[h]
	\centering
		\caption{\small Results for GTA5 (GTA)-to-Cityscapes (CS), SYNTHIA (SYN)-to-Cityscapes (CS) and Cityscapes (CS)-to-CrossCity (Rome, Rio, Tokyo, Taipei) experiments.}\vspace{-8pt}
	\resizebox{0.48\textwidth}{!}{    
{
			\begin{tabular}{l| c | c c c c c c }
				\toprule
				\multicolumn{8}{c}{Source Dataset $\to$ Target Dataset} \\
				\midrule
				Method & Backbone &
				\rotatebox{90}{GTA $\to$ CS} & \rotatebox{90}{SYN $\to$ CS} &
				\rotatebox{90}{CS $\to$ Rome} & \rotatebox{90}{CS $\to$ Rio} & \rotatebox{90}{CS $\to$ Tokyo} & \rotatebox{90}{CS $\to$ Taipei}  \\
				\hline
				Trans. Cons. w/o  $\mathcal{L}_p(o^{u},\tilde{o}^{u})$      
				& DeepLab-V2 &37.33 &33.48  &50.64      & 49.96      & 48.10         & 45.17         \\
				Trans. Cons. w/o $\mathcal{L}_g(\mathcal{A}_g(o^{u}),\hat{o}^{u})$
				& RN-101 &37.29 &33.51  &50.42      & 50.03      & 47.96        & 45.31          \\
				Trans. Cons.
				&  & {37.83} & {33.72}  & {50.82}      & {50.43}      & {48.20}         & {45.88}         \\
				\hline
				Trans. Cons. w/o $\mathcal{L}_p(o^{u},\tilde{o}^{u})$
                &  DeepLab-V2 &37.05 &32.84  & 51.62     & 49.60      & 47.44         & 47.38         \\
                Trans. Cons. w/o $\mathcal{L}_g(\mathcal{A}_g(o^{u}),\hat{o}^{u})$
                &  RN-101 &37.04 &32.97  & 51.65      & 49.58      & 47.57         & 47.29         \\ 
                Trans. Cons. 
                & + Transformer & {37.08} & {33.02}  & {51.67}      & {49.66}      & {47.91}         & {47.72}         \\
				\bottomrule    
	\end{tabular}}}

	\label{tab:photogeo}
\end{table}

\section{Visualization}
\begin{figure*}[h]
\centering
\includegraphics[width=0.98\linewidth]{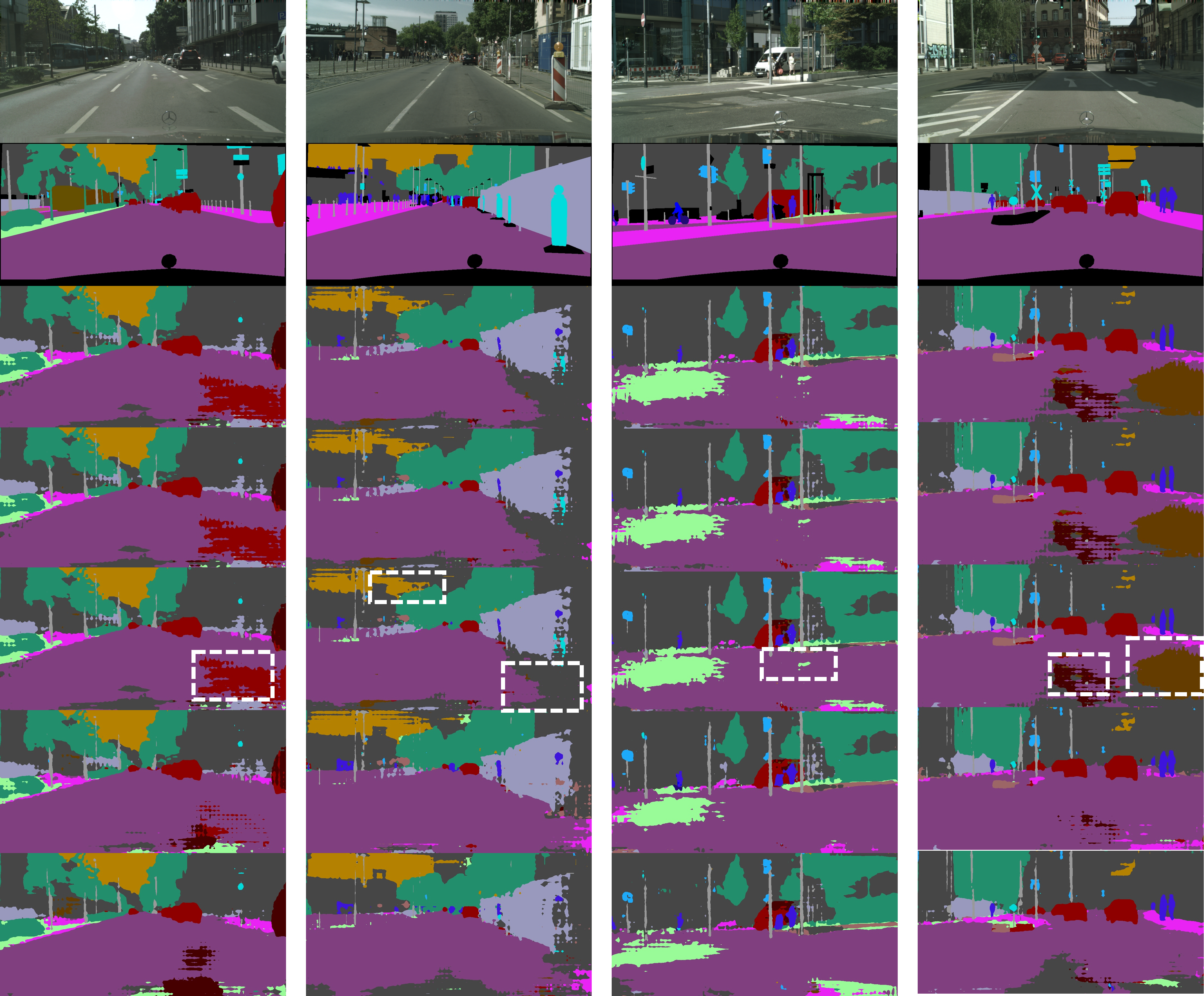}
\vspace{-5pt}
\caption{Segmentation map predictions for 4 different images from Cityscapes when  pre-trained on GTA5 and adapted on Cityscapes. From top to bottom: original image, ground truth, no-adaptation, max squares~\cite{ChenICCV19DASemSegmentationMaximumSquaresLoss}, min entropy~\cite{WangICLR21TENTFullyTestTimeAdaptEntropyMin}, transformation consistency and no adaptation with our transformer module. }
\label{fig:gta_viz}
\end{figure*}

\begin{figure*}[h]
\centering
\includegraphics[width=0.98\linewidth]{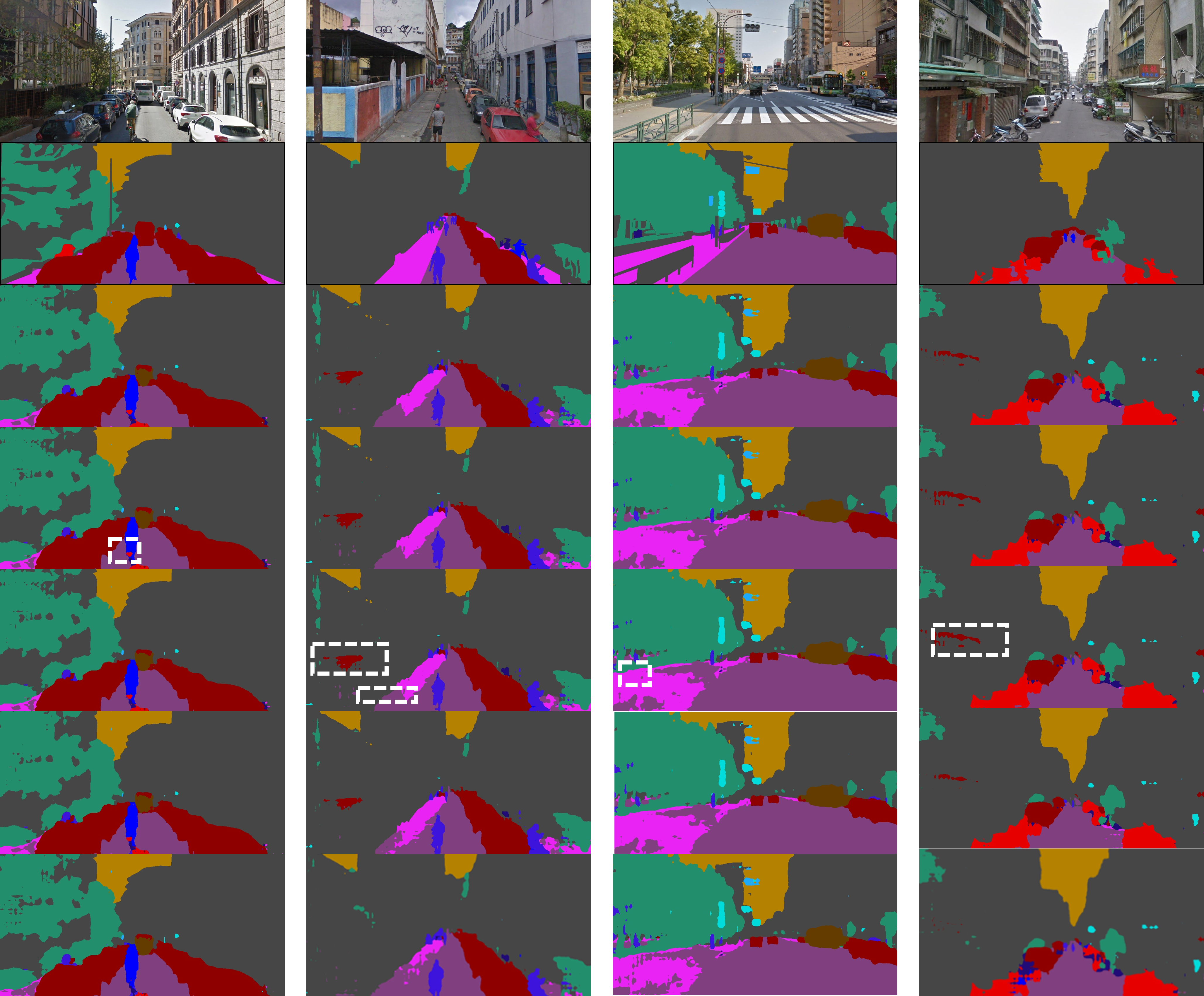}
\vspace{-5pt}
\caption{Segmentation map predictions for 4 different cities. From left to right: Rome, Rio, Tokyo, Taipei. Model has been pretrained on Cityscapes and adapted on each of the cities. From top to bottom: original image, ground truth, no-adaptation, max squares~\cite{ChenICCV19DASemSegmentationMaximumSquaresLoss}, min entropy~\cite{WangICLR21TENTFullyTestTimeAdaptEntropyMin}, transformation consistency and no adaptation with our transformer module. }
\label{fig:cs_viz}
\end{figure*}

We also visualize segmentation maps for two cross dataset setups: GTA5 $\to$ Cityscapes, Cityscapes $\to$ Cross-city. Results are shown in Figs.~\ref{fig:gta_viz} and~\ref{fig:cs_viz}, respectively. Overall, our proposed approaches of using transformation consistency and transformer module produces comparatively better segmentation maps. However, the improvement is less in the case of real-to-real domain shift of Cityscapes $\to$ Cross-city benchmark.

\end{document}